\documentclass{bmvc2k}
\usepackage{amssymb}
\usepackage{dsfont}
\usepackage{capt-of}

\title{JointFlow: Temporal Flow Fields for Multi Person Pose Tracking}

\addauthor{Andreas Doering}{doering@iai.uni-bonn.de}{1}
\addauthor{Umar Iqbal}{uiqbal@iai.uni-bonn.de}{1}
\addauthor{Juergen Gall}{gall@iai.uni-bonn.de}{1}

\addinstitution{
Computer Vision Group\\
 University of Bonn\\
 Bonn, DE
}

\runninghead{Doering, Iqbal, Gall}{JointFlow}

\def\eg{\emph{e.g}\bmvaOneDot}

\def\etal{\emph{et al}\bmvaOneDot}

\newcommand{\myNorm}[1]{\left\lVert#1\right\rVert}
\newcommand{\mySum}[2]{%
    \sum\limits_{#1}^#2%
}%

\newcommand{\myFigureHere}[5][image]%
{%
\begin{figure}[h]%
	\begin{center}%
		\includegraphics[width=#5\textwidth]{#2}%
		\caption[#4]{#3}%
		\label{#1_#2}%
	\end{center}%
\end{figure}%
}%

\newcommand{\myFigureTop}[5][image]%
{%
\begin{figure}[t]%
	\begin{center}%
		\includegraphics[width=#5\textwidth]{#2}%
		\caption[#4]{#3}%
		\label{#1_#2}%
	\end{center}%
\end{figure}%
}%

\begin{document}

\maketitle

\begin{abstract}
In this work we propose an online multi person pose tracking approach which works on two consecutive frames $I_{t-1}$ and $I_t$. The general formulation of our temporal network allows to rely on any multi person pose estimation approach as spatial network. From the spatial network we extract image features and pose features for both frames. These features serve as input for our temporal model that predicts Temporal Flow Fields (TFF). These TFF are vector fields which indicate the direction in which each body joint is going to move from frame $I_{t-1}$ to frame $I_t$. This novel representation allows to formulate a similarity measure of detected joints. These similarities are used as binary potentials in a bipartite graph optimization problem in order to perform tracking of multiple poses. We show that these TFF can be learned by a relative small CNN network whilst achieving state-of-the-art multi person pose tracking results.
\end{abstract}

\section{Introduction}
\label{sec:intro}

Understanding of human body pose is an important information for many scene understanding problems such as activity recognition, surveillance and human-computer interaction. Estimating the pose in unconstrained environments with multiple interacting people is a challenging problem. Apart from the large amounts of appearance variation and complex human body articulation, it also poses additional challenges such as large scale variation within a single scene, varying number of persons and body part occlusion and truncation. Multi-person pose estimation in videos increases the complexity even further since it also requires to tackle the problems of person association over time, large person or camera motion, motion blur, etc. 

In this work, we address the problem of multi-person pose tracking in videos, i.e, our goal is to estimate the pose of all persons appearing in the video and assign a unique identity to each person over time. The state-of-the-art approaches~\cite{detectNTrack,PoseFlow} in this direction build on the recent progress in multi-person pose estimation in images and first estimate the poses from images using off-the-shelf methods followed by an additional step for person association over time. There exist two main approaches for person association. The \textit{online} approach performs matching of the poses estimated at each time frame with the previously tracked poses and assigns an identity to each pose before moving to the next time step. In contrast, \textit{offline} or batch-processing based approaches~\cite{Iqbal_2017_CVPR} first estimate the poses in the entire video and then perform pose tracking while enforcing global temporal coherency of the tracks. In any case, both types of approaches require some metrics to measure the similarity between a pair of poses. The choice of the metrics and features used for matching plays a crucial role in the performance of these approaches. Recent methods for pose tracking~\cite{detectNTrack} rely on non-parametric metrics such as head normalized Percentage of Correct Keypoints (PCKh)~\cite{detectNTrack} and Object Keypoint Similarity (OKS)~\cite{FlowTrack} between a pair of poses, Intersection over Union (IoU) between the bounding boxes tightly enclosing each body pose~\cite{detectNTrack}, similarity between the image features extracted from the person bounding boxes~\cite{detectNTrack} or the optical flow information~\cite{Iqbal_2017_CVPR,arttrack,PoseFlow}. The location based metrics such as PCKh, OKS or IoU, on one hand, assume that the poses change smoothly over time, and therefore, struggle in case of large camera or body pose motion and scale variations due to camera zoom. On the other hand, appearance based similarity metrics or optical flow information  cannot handle large appearance variations due to person occlusions or truncation, motion blur, etc. The \textit{offline} approaches try to tackle these challenges by enforcing long-range temporal coherence. This is often done by formulating the problem using complex spatio-temporal graphs~\cite{Iqbal_2017_CVPR,arttrack} which results in very high inference time, and therefore, makes these methods infeasible for many applications. 

In this work, we present an approach for online multi-person pose tracking. In contrast to existing methods that rely on task-agnostic similarity metrics, we propose a task-specific novel representation for person association over time. We refer to this representation as Temporal Flow Fields (TFF). TFF represent the movement of each body part between two consecutive frames using a set of 2D vectors encoded in an image. Our TFF representation is inspired by the Part Affinity Fields representation~\cite{cao2017realtime} that measures the spatial association between different body parts and is learned by a CNN. We integrate TFF in an online multi-person tracking approach and demonstrate that a greedy matching approach is sufficient to obtain state-of-the-art multi-person pose tracking results on the PoseTrack benchmark~\cite{PoseTrack}. 
 
\section{Related Work}
The problem of multi person pose estimation in images has seen a drastic improvement over the last few years. Early works towards the direction of multi person pose estimation \cite{pishchulin12cvpr,
Eichner2010,
7360209,
DBLP:conf/cvpr/ChenY15, 
DBLP:conf/cvpr/LadickyTZ13} incorporate person detectors and estimate the corresponding poses based on  learning approaches (e.g.~random forests \cite{decision_forests}) combined with the pictorial structure model \cite{PSM}. 
With the introduction of deep learning based models, recent approaches achieve impressive multi person pose estimation results in images. These works can be divided into top-down \cite{Iqbal2016, mask_rcnn, fang2017rmpe, DBLP:conf/cvpr/PapandreouZKTTB17, DBLP:conf/cvpr/XiaWCY17, DBLP:journals/corr/abs-1711-07319} and bottom-up approaches \cite{deepcut16cvpr, deepercut, DBLP:journals/corr/NieFXY17, cao2017realtime, rogez:hal-01505085, DBLP:journals/corr/abs-1708-09182}. 

Former incorporate person detectors and estimate the pose for each person proposal.
For instance, Fang \etal \cite{fang2017rmpe} extend a stacked hourglass network \cite{stackedhourglass} by two transformer networks, a spatial transformer network (STN) and a spatial de-transformer network (SDTN) respectively. Each person proposal is passed to the STN which automatically detects the person of interest and applies an affine transformation which centers the person in an upright position. After pose estimation the SDTN maps the pose back to the input image. In a final step, the best pose for each person proposal is selected by non-maxima suppression on poses. 
Chen \etal \cite{DBLP:journals/corr/abs-1711-07319} categorize invisible or occluded keypoints as ``hard'' whereas the remaining keypoints are classified as ``simple''. The proposed cascaded model reflects their categorization of keypoints and is divided into two stages. The first stage is a feature pyramid network which detects ``simple'' keypoints (\eg head). The second stage, called \textit{RefineNet}, integrates all feature representations of different scales generated by the first stages. In that way, the \textit{RefineNet} is able to incorporate enough context to detect occluded or invisible body parts.   

Bottom-up approaches estimate the keypoints of all persons in a single run, but require a post-processing procedure to assemble these keypoints into person estimates. 
Cao \etal \cite{cao2017realtime} estimate multiple poses in real-time. Their work extends the model proposed in \cite{wei2016cpm} by introducing an additional branch which predicts vector fields between body parts of individual persons. These so called Part Affinity Fields (PAF) preserve location and orientation information \cite{cao2017realtime} of limbs. The authors propose to use a greedy bipartite graph matching algorithm, which greedily connects joints that share the same body part.
The work of Varadarajan \etal \cite{DBLP:journals/corr/abs-1708-09182} introduces a more efficient greedy part assignment algorithm compared to \cite{cao2017realtime}. 
After part belief maps and pairwise association maps are obtained like in \cite{cao2017realtime}, the number of part candidates is reduced to an approximate number of persons within a clustering step. By following the kinematic chain, body parts are assigned in a greedy fashion to joints of the most proximal candidate person cluster.

\subsection{Multi-Person Pose Tracking}
Even though a big advancement in multi person pose estimation in images has been achieved, very few works have addressed this problem in videos \cite{Iqbal_2017_CVPR, arttrack, detectNTrack, FlowTrack}. \cite{Iqbal_2017_CVPR} is one of the first works which tackels the problem of multi person pose estimation and tracking by solving a spatio-temporal graph matching problem. The spatio-temporal graph is created by densely connecting all detected joint candidates in the spatial domain. In the temporal domain all joints of the same class are connected. In order to find the best graph partition, a conditioned integer linear programming problem has to be optimized. For runtime reasons, \cite{Iqbal_2017_CVPR} propose to sequentially optimize for temporal windows of a fixed size only. Nevertheless, the runtime is still too high which makes this work impractical for real-time applications. A very similar approach with comparable performance is proposed by \cite{arttrack} which in contrast to \cite{Iqbal_2017_CVPR} relies on a sparse spatio-temporal graph.  \cite{detectNTrack} propose a video pose estimation formulation which consists of a 3D extension of the \textit{Mask R-CNN} model \cite{mask_rcnn}. By integrating temporal information, the proposed model estimates person bounding boxes and poses which the authors refer to as \textit{person tubes}. To achieve this, their network first predicts bounding boxes for each frame followed by a pre-trained Resnet-101 network  \cite{resnet} for pose estimation. In order to link the estimated poses in time, \cite{detectNTrack} propose to solve a bipartite graph matching problem in a greedy fashion and show that the achieved results are very close to the optimal solution obtained via Hungarian algorithm. By comparing different distance metrics, the authors show that Intersection over Union (IoU) of person bounding boxes achieves the best trade-off between performance and runtime. Nevertheless, this approach requires to process entire sequences or portion of a sequence which limits the applicability for real-time applications. 

In \cite{FlowTrack}, the authors follow a very similar baseline as proposed in \cite{detectNTrack}, but in contrast the authors rely on two different sources for person bounding boxes: a bounding box detector and optical flow. This allows to warp estimated poses of the previous frames $I_{\Delta t}$ with $\Delta t = \lbrace 1, \ldots T \rbrace$ into the current frame and a similarity metric between estimated and warped poses based on the Object Keypoint Similarity (OKS) is used for the calculation of binary potentials of a temporal graph. By utilizing greedy graph matching similar to \cite{detectNTrack} this approach achieves state-of-the-art results.

\section{Overview}

\myFigureTop[pipeline]{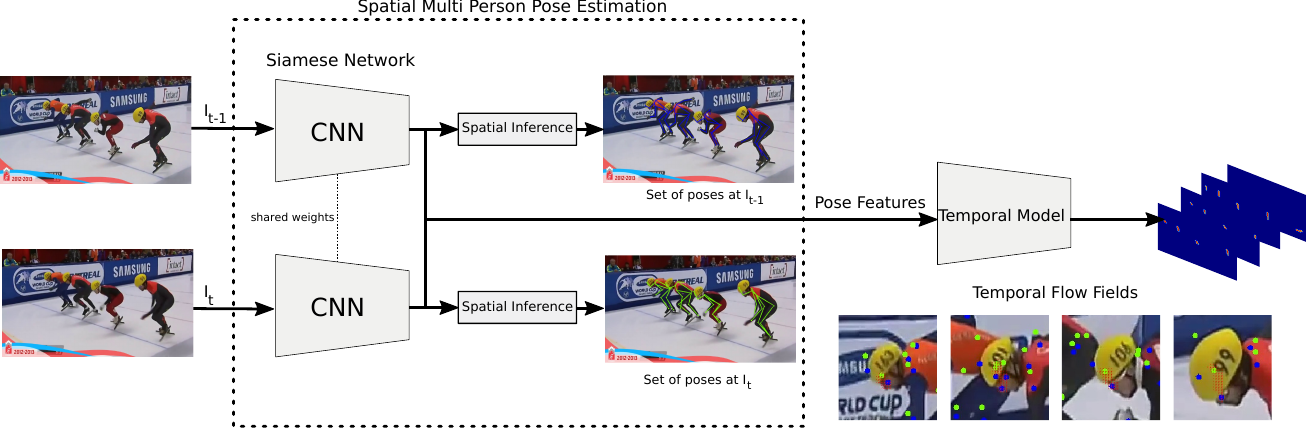}{Proposed approach: For two consecutive input frames $I_{t-1}$ and $ I_t$, we utilize a Siamese network initialized by an arbitrary multi person pose estimation network. During spatial inference, pose features such as belief maps or part affinity fields are used to estimate the poses for each frame. Building on these pose features, our porposed temporal model predicts the temporal flow field for each detected joint which are used during inference to associate poses in time.}{}{1}
In this work, we propose to predict Temporal Flow Fields in an online fashion. To this extend, we evaluate two frames at a time as visualized in Figure \ref{pipeline_images/chapter3/bmvc_pipeline.pdf} and estimate their poses. The structure of our temporal model allows to utilize any network architecture for the task of multi person pose estimation. In the context of this work, we use the CNN of \cite{cao2017realtime} as a component in our Siamese network. 
While the Siamese network is used to predict the poses in both frames, we take the last layer as input for the temporal CNN which predicts the Temporal Flow Fields (TFF). 
To track the poses, we  then create a bipartite graph $\mathcal{G}$ as illustrated in Figure \ref{fig:temporal_graphs}b) from the estimated poses and use the estimated TFF as similarity measure (Sec. \ref{sec:mppt}) in a bipartite graph matching problem.

\section{Multi-Person Pose Tracking}\label{sec:mppt}
We represent the body pose $P$ of a person with $J$ body joints as $P={\mathbf\{p}_j\}_{1:J}$, where \mbox{$p_j=(x_j,y_j)$} represent the 2D pixel coordinates of the $j^{th}$ body joint. Given an input video, our goal is to perform multi person pose estimation and tracking in an online manner. Formally, at every time instance $t$ with video frame $I_t$ containing $N_t$ persons, we first estimate a set of poses $\mathcal{P}_{t} = \lbrace {P}_{t}^1, \ldots {P}_{t}^{N_t} \rbrace$ and then perform person association with the set of persons  $\mathcal{P}_{t-1} = \lbrace {P}_{t-1}^1, \ldots {P}_{t-1}^{N_{t-1}} \rbrace$ tracked until the last video frame $I_{t-1}$. For pose estimation, we use an improved version of \cite{cao2017realtime} that we will explain briefly in Sec.~\ref{sec:implementation_details}. We formulate the problem of person association between the set of poses $\mathcal{P}_t$ and $\mathcal{P}_{t-1}$ as an energy maximization problem over a bipartite graph $\mathcal{G}$ (Figure \ref{fig:temporal_graphs}b) as follows

\begin{eqnarray}\label{eqt:objective}
    \hat{z} & = & \underset{z}{\textit{argmax}} \mySum{P_t \in \mathcal{P}_{t}}{\text{ }} \mySum{P_{t-1}' \in \mathcal{P}_{t-1}}{\text{ }}  \Psi_{P_{t},P_{t-1}'} \cdot z_{P_{t},P_{t-1}'}\\
     \text{s.t.} & & \forall P_t \in \mathcal{P}_{t}, \; \mySum{P_{t-1}' \in \mathcal{P}_{t-1}}{\text{ }} z_{P_{t}, P_{t-1}'} \leq 1 
     \;\;\; \text{and} \;\;\;\forall P_{t-1}' \in \mathcal{P}_{t-1}, \; \mySum{P_t \in \mathcal{P}}{\text{ }} z_{P_t, P_{t-1}'} \leq 1, \nonumber
\end{eqnarray}
where $z_{P_t,P_{t-1}'} \in \{0,1\}$ is a binary variable which indicates that the poses $P_t \in \mathcal{P}_t$ and $P_{t-1}' \in \mathcal{P}_{t-1}$ are associated with each other, and the binary potentials $\Psi_{P_t,P_{t-1}'}$ define the similarity between the pair of poses $P_t$ and $P_{t-1}'$. 

\subsection{Temporal Flow Fields}
\myFigureTop[tpaf]{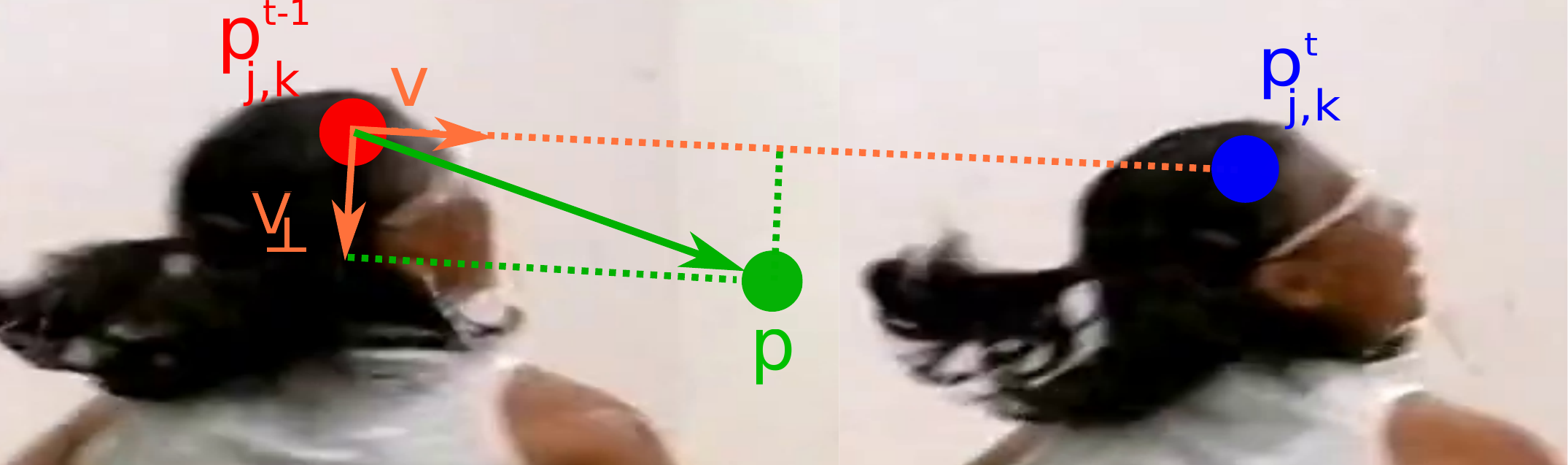}{Calculation of Temporal Flow Fields: Let $p_{j,k}^{t-1}$ and $p_{j,k}^t$ be the location of joint $j$ of person $k$ in frames $I_{t-1}$ and $I_t$. For every point $p \in \Omega_{j,k}$ located on the flow field, the TFF $T_{j,k}^*(p)$ contains a unit vector v and 0 otherwise.}{}{0.8}

We model the binary potentials $\Psi_{P_t,P_{t-1}'}$ \eqref{eqt:objective} by Temporal Flow Fields (TFF) and define each TFF as a vector field that contains a unit vector $v$ for each pixel $p = (x, y)$.  Each unit vector $v = \frac{p_{j, k}^t - p_{j, k}^{t-1}}{\lambda_{j,k}}$ points towards the direction of the target joint location $p_{j, k}^t \in P_t^k$ where \mbox{$\lambda_{j,k} = ||p_{j, k}^t - p_{j, k}^{t-1}||_2$}  is the Euclidean distance between the estimated joint locations of person $k$ in frames $I_{t-1}$ and $I_t$. We restrict the TFF to pixels that are close to the joint motion by a parameter $\sigma$ and describe the set of pixels of the TFF as
\begin{equation}\label{3_2:omega_t}
    \Omega_{j,k} = \lbrace p \; \vert \; 0 \leq v \cdot (p - p_{j,k}^{t-1}) \leq  \lambda_{j,k} \land \vert v_{\bot} \cdot (p - p_{j, k}^{t-1}) \vert \leq \sigma \rbrace,
\end{equation}
where $v_{\bot}$ is a unit vector perpendicular to $v$ as illustrated in Figure \ref{tpaf_images/chapter3/tpaf_visualization}. This allows a pixel-wise definition of TFF for joint class $j$ of person $k$ 

\begin{eqnarray}\label{3_2:temporal_paf_eq}
    T_{j,k}^{\text{*}}(p) &=& \left\lbrace \begin{tabular}{cc}
        $v$ & \text{if} $p \in \Omega_{j,k}$ \\
        0 & otherwise.
    \end{tabular}\right.
\end{eqnarray}
In a final step, a single representation of a flow field $T_j$ is generated for each joint class by aggregating the TFF among all estimated persons. 
\begin{equation}
    T^*_{j}(p) = \frac{1}{n_t(p)} \mySum{k=1}{{K}} T^*_{j,k}(p),
\end{equation}
where $ n_t(p)$ is the number of non-zero unit vectors $v$ at location $p$ across all $K$ persons.
\subsubsection{Model}
\myFigureTop[siamese]{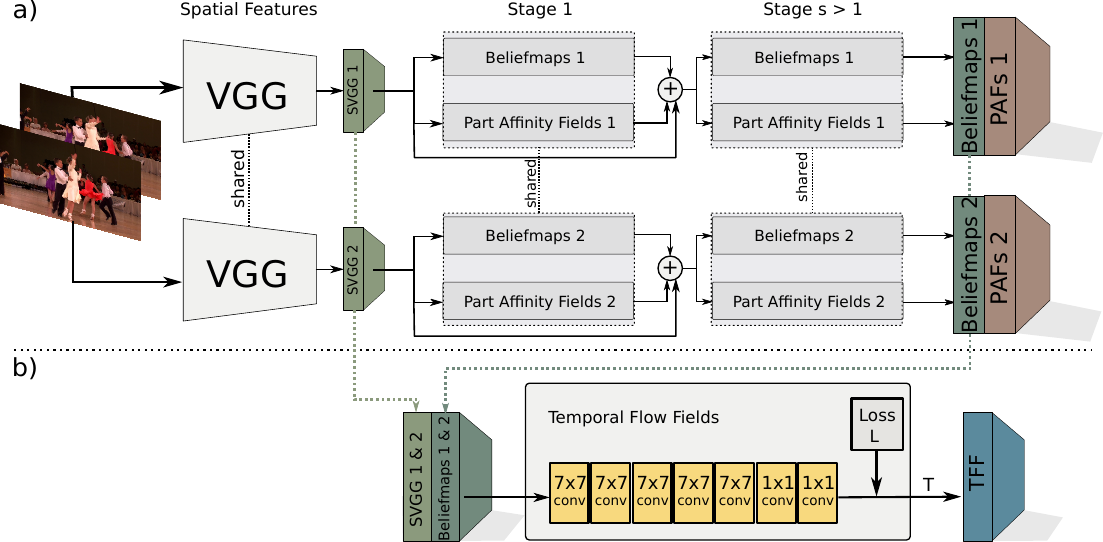}{Proposed Model Structure: a) Siamese network to extract pose features (\textit{SVGG, Belief, PAF}) for frames $I_{t-1}$ and $ I_t$ and b) temporal network to extract temporal part affinity fields for feature map input SVGG and Beliefmaps (\textit{SVGG + Belief}) from the two frames.}{Proposed Model Structure}{0.7}

For the prediction of Temporal Flow Fields, we propose an efficient CNN as illustrated in Figure \ref{siamese_images/chapter3/siamese_shrinked}b) which consists of five $7\times 7$-convolution layers with a stride of one pixel followed by two $1\times 1$-convolutions. Non-linearity is achieved by ReLU layers after each convolution.

As input, the network expects image features and pose features. These are obtained from the Siamese network visualized in Figure \ref{siamese_images/chapter3/siamese_shrinked}a) which is initialized by a modified version of \cite{cao2017realtime} consisting of six stages.  In particular, image features of both frames $I_{t-1}$ and $I_t$ are obtained by a feature extraction layer as illustrated in Figure \ref{siamese_images/chapter3/siamese_shrinked}a). We refer to these image features as \textit{SVGG}. Additionally, the Siamese network predicts beliefmaps and Part Affinity Fields (PAFs) (cf. \cite{cao2017realtime}) at each stage which we refer to as pose features.
Based on image features and pose features, extracted from the last stage, the temporal model predicts TFF.  

For the training of our model, we calculate the weighted squared L2 loss of the form 
\begin{equation}\label{eq:loss}
    \mathcal{L} = \mySum{j=1}{J}\mySum{p \in \Omega}{\text{}} M(p) \cdot \myNorm{T_j^*(p) - T_j(p)}_2^2,
\end{equation}
where $T_j^*(p)$ and $T_j(p)$ are the ground truth TFF and the predicted TFF at pixel location $p$ respectively. $M$ is a binary mask with $M(p)=0$ for all pixels located on an ignore region, i.e., a region for which the dataset does not provide any annotations. 

\subsection{Inference}
\begin{figure}[t]
\begin{center}
\includegraphics[width=0.9\textwidth]{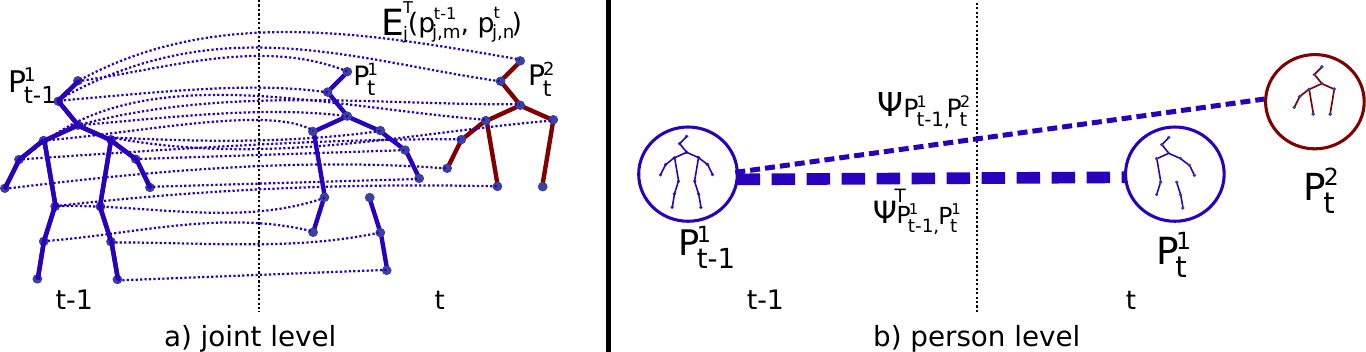}
\end{center}
\caption{Temporal edge candidate generation for incomplete pose estimates: a) on a joint level and b) on a person level where $\Psi_{P_{t-1}^1, P_t^1}$ and $\Psi_{P_{t-1}^1, P_t^2}$ are the accumulated temporal edge potentials estimated according to Equation (\ref{3_4:person-level-aggregate}).}
\label{fig:temporal_graphs}
\end{figure}

During inference, we partition the bipartite graph $\mathcal{G}$ by optimizing \eqref{eqt:objective} using a greedy approach. In order to obtain the binary potentials $\Psi_{P_{t-1}, P_t'}$ we first generate $J$ temporal bipartie subgraphs $\mathcal{G}_j$ with a set of edges
    $Z_{j}^t  =  \left\lbrace \left.  z_j^{p_{t-1},p_{t}} \right\vert  P_{t} \in \mathcal{P}_t , P_{t-1} \in \mathcal{P}_{t-1}  \right\rbrace$
that connect all detected joints of class $j$ in frame $I_{t-1}$ with all detected joints in frame $I_t$ of the same class. Figure \ref{fig:temporal_graphs}a) illustrates such subgraphs. Along each temporal edge of $\mathcal{G}_j$, we follow the estimated TFF and obtain a flow field aggregate given by 
\begin{equation}\label{3_4:tpaf_energy}
    E_{aggr}(p^{t-1}_{j, m}, p^t_{j, n}) = \int_{o=0}^{o=1} T_{j}(i(o))^{\top} \frac{(p^t_{j, n} - p^{t-1}_{j, m})}{\myNorm{p^t_{j, n} - p^{t-1}_{j, m}}_2}do,
\end{equation}
where $i(o) = (1-o) \cdot p_{j, m}^{t-1} + o \cdot p_{j, n}^t$  is a function that interpolates the location between both detected joints  $p_{j, m}^{t-1}$ and $p_{j, n}^t$. The value is high if the TFF points in the same direction as $p_{j,n}^t - p_{j,m}^{t-1}$ along $i(o)$. In addition to this formulation, we have to consider a special case: if there is no motion of a joint between frames, no flow field would exist and the flow field aggregate would be zero. To overcome this issue, we incorporate the Euclidean distance $\Delta p_{j, m, n}^t = ||p_{j,n}^t - p_{j, m}^{t-1}||_2$ between both joint locations into our similarity measure and define 

\begin{equation}\label{3_4:temporal_joint_potential_eq}
  E_j(p_{j,m}^{t-1}, p_{j,n}^{t}) = \left\lbrace \begin{tabular}{ll}
        $ E^T_{aggr}(p_{j,m}^{t-1}, p_{j,n}^t)$ & \text{if} $\Delta p_{j,m,n}^t \geq \tau_{\Delta}$ \\
        1 &  \text{if} $\Delta p_{j,m,n}^t < \tau_{\Delta}$,
    \end{tabular} \right. 
\end{equation}
where $\tau_{\Delta}$ is a pre-defined distance threshold. This definition allows to formulate the binary potentials required to solve \eqref{eqt:objective}. 
However, instead of solving $J$ different bipartite graph matching problems for each joint class $j$, we convert each estimated pose into a node of graph $\mathcal{G}$ as illustrated in Figure \ref{fig:temporal_graphs}b). The temporal potential $\Psi_{P_{t-1}^{m}, P_t^{n}}$ is then defined as the accumulated similarity between all joints $j$ of persons $P_{t-1}^m$ and $P_t^n$ 

\begin{equation}\label{3_4:person-level-aggregate}
    \Psi_{P_{t-1}^m,P
    _t^n} = \mySum{j=1}{J} \mathds{1}(p_{j, m}^{t-1}, p_{j,n}^t) \cdot E_j(p_{j,m}^{t-1}, p_{j, n}^{t}),
\end{equation}
where $\mathds{1}(p_{j, m}^{t-1}, p_{j,n}^t)$ is a binary function with $\mathds{1}(p_{j, m}^{t-1}, p_{j,n}^t) = 1$ if both joints are detected.
An example is shown in Figure \ref{fig:temporal_graphs}a) and \ref{fig:temporal_graphs}b): temporal edges between estimated joints of person $P^1_{t-1} \in \mathcal{P}_{t-1}$ of frame $t-1$ and persons $P^1_t, P^2_t \in \mathcal{P}_t$ in frame $t$ are estimated. In this particular case, persons $P^1_{t-1}$ and $P^1_t$ share 12 temporal connections among joints whereas persons $P^1_{t-1}$ and $P^2_{t}$ share 9 temporal edges. The costs of these edges are accumulated and assigned as potential $\Psi_{P^1_{t-1},P^1_{t}}$ and $\Psi_{P^1_{t-1},P^2_{t}}$ to the temporal connections among persons as shown in Figure \ref{fig:temporal_graphs}b).

By solving \eqref{eqt:objective}, we assign each detected person to either one of the poses from the previous frame, which continues the track, or a new track is initialized if no assignment is possible.

\section{Implementation Details}\label{sec:implementation_details}

\noindent
For the spatial part of our proposed approach (Figure \ref{pipeline_images/chapter3/bmvc_pipeline.pdf}), we re-implemented the method of \cite{cao2017realtime} and applied two minor modifications: instead of initializing the feature extraction part by 10 layers of VGG19, we increase the number of layers to 12. The second modification includes a different edge configuration for the prediction of Part Affinity Fields. Both changes result in a gain in pose estimation performance. For further evaluations and a detailed description of the underlying edge configuration, we refer to the supplementary material. For the detection of joints, we perform Non-Maximum Suppression (NMS) on the estimated beliefmaps and discard all detections that do not meet a threshold $\tau_{NMS} = 0.2$.
The spatial model was trained on the MSCOCO dataset \cite{Lin2014} for 22 epochs with a learning rate of $\eta=4\cdot 10^{-5}$ and a decay in learning rate of $\gamma=0.333$ after 7 epochs. For finetuning, we rely on the PoseTrack dataset \cite{PoseTrack} and train for 3000 iterations with a learning rate of $\eta = 10^{-5}$ and a decay in learning rate after 1000 iterations.

Our temporal model is trained on the PoseTrack dataset for 40 epochs with a learning rate $\eta=4\cdot 10^{-6}$ and a learning rate decay of $\gamma = 0.333$ after seven epochs. 
Additionally, we fix $\tau_{\Delta} = 2$ and select $\sigma = 1$ as the desired width of our Temporal Flow Fields.

During inference, we evaluate pairs of frames at four different scales (0.5, 1, 1.5 and 2) and average the estimated results. Spatial greedy bipartite graph matching is performed similar to \cite{cao2017realtime} to estimate the poses first, followed by our proposed pose tracking approach. Since we did not focus on optimizing the runtime as performed in \cite{cao2017realtime}, inference requires on average 5.6 seconds for a pair of images on an I7-5820K @ 3.3 GHz and a single 1080TI. Unlike \cite{cao2017realtime}, we did not resize the videos of the PoseTrack dataset, which also contains HD quality frames, to a smaller resolution which leaves a lot of space for improvements in runtime. 
\section{Experiments}\label{4:evaluation}

\begin{table}[t]
        \begin{minipage}[t]{0.5\textwidth}
            \centering
            	\scalebox{0.6}{\begin{tabular}{llllll}
			\hline\noalign{\smallskip}
			 & & MOTA & MOTP & Prec & Rec  \\
			\noalign{\smallskip}
			\hline
			\noalign{\smallskip}
			Input Features & $\tau_{NMS}$ & Total& Total& Total& Total\\
			\noalign{\smallskip}
			\hline
			\noalign{\smallskip}
	        SVGG  & 0.1 & 56.0 & 84.2 & 82.4 & 74.9 \\
	        \textbf{SVGG + Belief} & \textbf{0.1} &  \textbf{56.2} & \textbf{84.2} & \textbf{82.4} & \textbf{74.9} \\
	        SVGG + Belief + PAF  & 0.1 &  \textbf{56.2} & 84.2 & 82.4 & 74.9 \\
			\noalign{\smallskip}
			\hline
		\end{tabular}}
\caption{Impact of different combinations of input features on the pose tracking performance.}\label{table:diff_input_features}
        \end{minipage}
        \hfill
        \begin{minipage}[t]{0.5\textwidth}
            \centering
            	\scalebox{0.6}{\begin{tabular}{lllllll}
			\hline\noalign{\smallskip}
		    & & MOTA & MOTP & Prec & Rec & mAP\\
			\noalign{\smallskip}
			\hline
				\noalign{\smallskip}
			Input Features & $\tau_{NMS}$ & Total & Total & Total & Total & Total\\
		    \noalign{\smallskip}
			\hline
			\noalign{\smallskip}
            SVGG + Belief & 0.1 & 56.2 & 84.2 &  82.4 & 74.9 & 69.3 \\
             \textbf{SVGG + Belief} & \textbf{0.2} & \textbf{59.1} & \textbf{84.4} & \textbf{87.1} & \textbf{71.9} & \textbf{67.0} \\
            SVGG + Belief & 0.3 & 58.2  & 84.9  & 91.1  & 66.8  & 62.4 \\
			\hline
		\end{tabular}}
\caption{Impact of threshold $\tau_{NMS}$ during NMS of beliefmaps on pose estimation and tracking performance.}\label{table:nms_threshold}
        \end{minipage}
    \end{table}
    
Within the first experiment, we tested different combinations of pose features (Table \ref{table:diff_input_features}) in order to evaluate their performance. To this extend, each model was trained for 40 epochs on the PoseTrack dataset \cite{PoseTrack} and we rely on the metrics proposed in \cite{MOTA} in order to measure pose tracking performance. Certainly, spatial image features (SVGG) provide a strong cue for the prediction of temporal vector fields of each joint class.  Additional knowledge is provided by the estimated belief maps (Belief) of both input frames. Part Affinity Fields (PAFs) \cite{cao2017realtime} represent the skeletonal structure of the human body, and were expected to boost the performance even further. As Table \ref{table:diff_input_features} reveals, this is not the case. In order to reduce the number of input parameters, we use \textit{SVGG + Belief} as desired input for the temporal model. In order to evaluate the impact of an increased receptive field, we explored the impact of multiple stages similar to the spatial model. Our experiments have shown that further stages do not have any impact on the final performance.  

In an additional set of experiments, we evaluate different thresholds $\tau_{NMS}$ for non-maximum suppression of the heatmaps used for the detection of joint candidates. Even though a higher threshold results in less accurate pose estimates, more confident detection candidates result in stronger person tracks. According to Table \ref{table:nms_threshold}, we select $\tau_{NMS}=2$ as passable trade-off resulting in a boost in tracking performance.

\subsection{Comparison to Baselines}	
In an additional set of experiments, the performance of TFF is compared to different tracking metrics, namely Intersection over Union (IoU) of persons, PCKh \cite{andriluka14cvpr}, Object Keypoint Similarity (OKS) and optical flow based tracking. For this purpose, the temporal potential defined in \eqref{3_4:person-level-aggregate} has to be adapted to
\begin{equation}\label{4:adapted_aggregate}
    \Psi_{P_{t-1}^m, p_t^n} = \left\lbrace \begin{tabular}{ll}
         IoU($BB_{P_{t-1}^m}, BB_{P_t^n}$) & \text{for IoU}  \\
         PCKh($P_{t-1}^m, P_t^n$) & \text{for PCKh} \\
         OKS($P_{t-1}^m, P_t^n$) & \text{for OKS},
    \end{tabular} \right.
\end{equation}
where $BB_{P_{t-1}^m}$ and $ BB_{P_t^n}$ are the bounding boxes for persons $P_{t-1}^m \in \mathcal{P}_{t-1}$ and $P_t^n \in \mathcal{P}_t$ in frames $I_{t-1}$ and $I_t$ respectively. The bounding boxes for each person are estimated from the detected poses. Table \ref{table:baselines} summarizes the results. All three metrics can not compete with the proposed TFF. 
\begin{table}[t]
    \centering
      \scalebox{0.8}{\begin{tabular}{lllll}
			\hline\noalign{\smallskip}
			 &  MOTA & MOTP & Prec & Rec  \\
			\noalign{\smallskip}
			\hline
			\noalign{\smallskip}
			 Baselines  & Total& Total& Total& Total\\
			\noalign{\smallskip}
			\hline
			\noalign{\smallskip}
	        PCKh                &   50.0    & 84.4  & 87.1  & 71.9 \\
            IoU                 &   57.7    & 84.4  & 87.1  & 71.9 \\ 
            OKS                 &   58.8    & 84.4  & 87.1 & 71.9 \\ 
            Optical Flow        &   58.5    & 84.4  & 87.1  & 71.9 \\
            \textbf{Temporal Flow Fields} &  \textbf{59.1} & \textbf{84.4} & \textbf{87.1} & \textbf{71.9} \\
           
			\hline
		\end{tabular}}
    \caption{Comparison to different baselines}
    \label{table:baselines}
    \label{tab:my_label}
\end{table}

Optical flow based tracking requires a different set of changes. First of all, we rely on the approach of \cite{IMKDB17} in order to estimate the optical flow $f \in \mathcal{R}^{w \times h \times 2}$.
Similar to Temporal Flow Fields, the optical flow is a vector field which can be used to predict the movement of each joint from frame $I_{t-1}$ to frame $I_t$. In order to incorporate the optical flow into the greedy bipartite graph matching algorithm, the flow field aggregation energy (\ref{3_4:tpaf_energy}) has to be adapted as follows:
\begin{equation}\label{flow_aggregate_eq}
    E_{flow}^T(p_{j,m}^{t-1}, p_{j, n}^{t}) = e^{-\frac{\myNorm{p_{j,n}^t - (p^{t-1}_{j, m} + f(p^{t-1}_{j, m}))}^2}{\sigma^2_{flow}}},
\end{equation}
where $\sigma_{flow}$ controls the tolerance radius to mistakes. In that way, optical flow vectors $f$ which vote for locations close to $p^t_{j, n}$ still contribute significantly to the energy $E_{flow}^T$. Experiments have shown, that $\sigma_{flow}=30$ performs best. 
Although the network for optical flow \cite{IMKDB17} is much larger and more expensive than our network for TFF, TFF outperform the optical flow.

\subsection{Comparison to State-of-the-Art}
For a comparison to the state-of-the-art, we compare to the results on the PoseTrack validation set reported in \cite{FlowTrack, PoseFlow, detectNTrack} and to the results on the PoseTrack test set taken from the PoseTrack challenge leaderboard \cite{leaderboard}.
On the validation set, we achieve a total MOTA of 59.1 which can be improved up to 59.8 after pruning tracks of a length smaller than 7 frames. 
We submitted our results to the official validation server and achieved the second place on the leaderboard with a final MOTA of 53.1. 

	\setlength{\tabcolsep}{4pt}
	\begin{table}
		\centering
		
		\label{table:state-of-the-art}
		
		\scalebox{0.75}{\begin{tabular}{llllll}
			\hline\noalign{\smallskip}
			 &  & MOTA                    & Prec      & Rec & mAP \\
			\noalign{\smallskip}
			\hline
			\noalign{\smallskip}
 Approach       & Evaluation Set    &Total                 & Total& Total & Total\\
			\noalign{\smallskip}
			\hline
			\noalign{\smallskip}
			FlowTrack  \cite{FlowTrack}             & val   & 65.4  &  85.5  & 80.3 & 76.7\\ 
			TFF                          & val   & 59.1 &   87.1 &  71.9 & 69.3 \\
            \textbf{TFF + pruning}             & val   & 59.8 &  87.8 & 71.1 & 66.7\\
    	    PoseFlow \cite{PoseFlow}            & val   & 58.3 & 87.0    & 70.3 &	66.5 \\
			ProTracker \cite{detectNTrack}          & val   & 55.2  & 88.1  & 66.5 & 60.6 \\
            \noalign{\smallskip}
			\hline
			\noalign{\smallskip}
			FlowTrack  \cite{FlowTrack}             & test  & 57.8       &  79.4     & 80.3 & 74.6 \\
			\textbf{TFF + pruning}             & test  & 53.1 & 82.6 & 69.7 & 63.3 \\
			HMPT$^*$                                    & test & 51.9 & - & - & 63.7 \\
			ProTracker \cite{detectNTrack}          & test  & 51.8          & -     & - & 59.6 \\
			PoseFlow \cite{PoseFlow}                & test  & 51.0      & 78.9     & 71.2 &	63.0 \\
            MVIG$^*$                                   & test  & 50.8            & -     & -  & 63.2 \\
            BUTD2$^*$                                   & test & 50.6           & -     & -     & 59.2 \\
            Trackend$^*$                                & test & 49.7           & -     & -     & 57.5 \\
            PoseTrack \cite{PoseTrack}                  & test & 48.4           & -     & -     & 59.4 \\
            MIPAL$^*$                                   & test & 46.3           & -     & -     & 69.9 \\
            SOPT-PT $^*$                                & test  & 42.0          & -     & - & 58.2\\
            ML\_Lab$^*$                                 & test  & 41.8          & -     & - & 70.3\\
            ICG$^*$                                     & test  & 32.0          & -     & - & 51.2\\
            IC\_IBUG$^*$                                & test  & -190.1        & -     & - & 47.6\\
            \hline
			
			\hline
		\end{tabular}}
		\caption{Comparison to state-of-the-art. Approaches marked with $^*$ have not been published yet.}
	\end{table}
	\setlength{\tabcolsep}{1.4pt}

\section{Conclusions}

In this work, we proposed a convolutional neural network architecture for the task of online multi person pose tracking. Our approach consists of two sub-networks: a spatial network for multi person pose estimation and a temporal network which predicts Temporal Flow Fields. TFF are used by a greedy temporal bipartite graph matching algorithm which associates estimated poses in two consecutive frames $I_{t-1}$ and  $I_t$. 
The results showed that a strong structural knowledge in form of image features and belief maps of both frames are crucial for a good performance of our temporal model. By relying on such feature input, our approach achieves state-of-the-art pose tracking results, even with a small network architecture. For this reason, in future work we will investigate stronger network architectures in order to produce stronger Temporal Flow Fields which are able to cope with additional challenges like occlusions and long-term dependencies. 

\section{Acknowledgments }
The work has been financially supported by the DFG projects
GA 1927/5-1 (DFG Research Unit FOR 2535 Anticipating Human Behavior) and the ERC Starting Grant ARCA (677650).

\bibliography{egbib}
\newpage
\appendix
\section{Supplementary Material}
\subsection{Qualitative Results}
\myFigureHere[subFig]{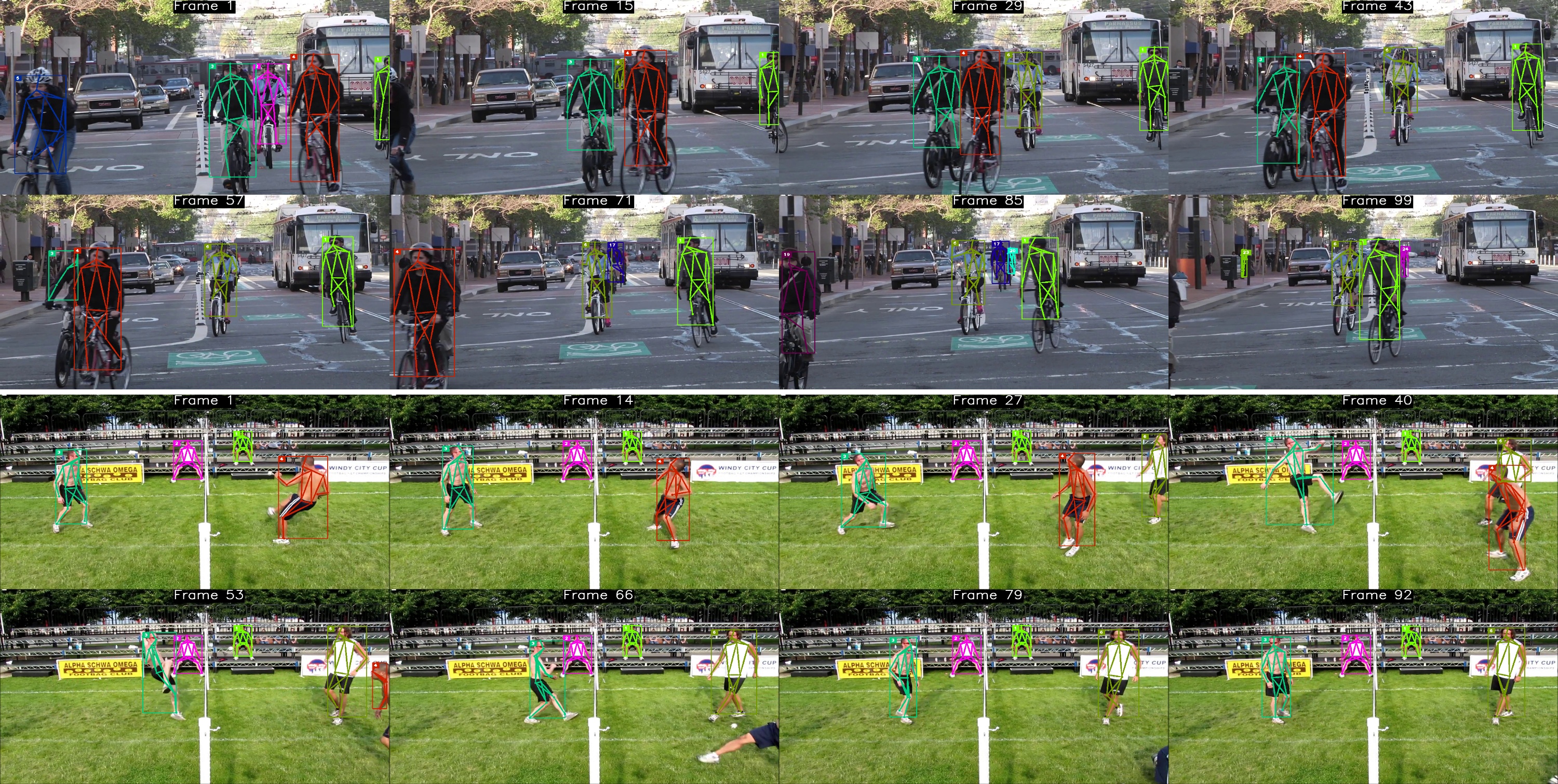}{Qualitative results for sequences of the PoseTrack validation set \cite{PoseTrack}.}{Qualitative Results}{1}
	
\subsection{Baseline Improvement}\label{4:baseline_improvement}
\myFigureHere[exp]{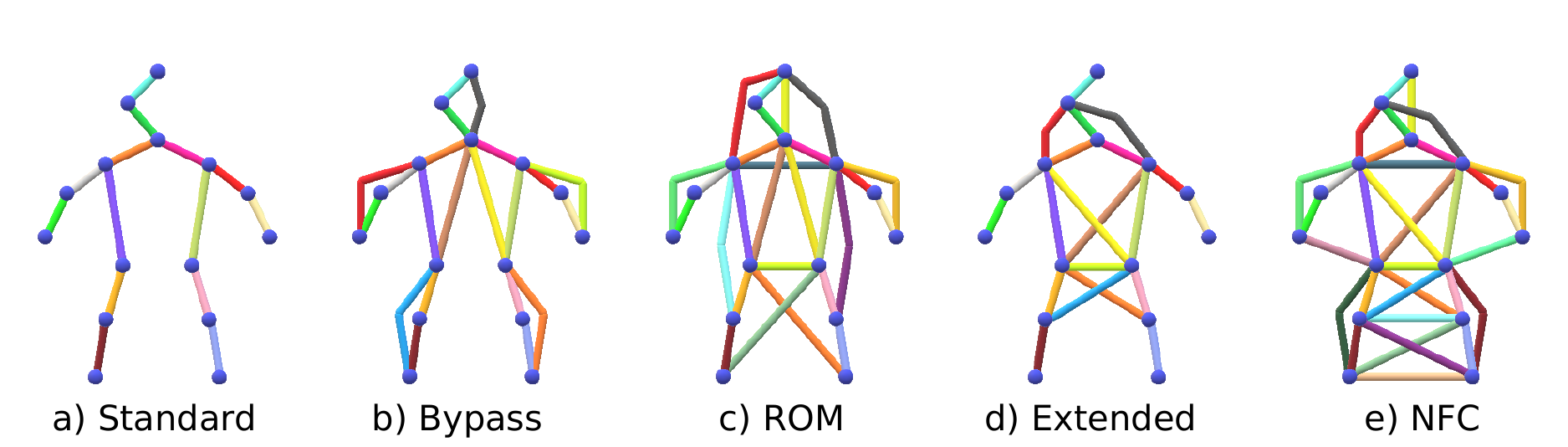}{Different edge configurations used for the training of different spatial models.}{Different Edge Configurations during Training}{1}
We evaluate the robustness of different edge configurations as shown in Figure \ref{exp_images/chapter5/edges_experiments}. This is motivated by the fact that edge configuration a) is prone to errors. If a single edge is not estimated correctly, the entire pose breaks. Similar to \cite{samsung} we introduce skip connections to the standard model (Figure \ref{exp_images/chapter5/edges_experiments} b) \textit{Bypass model}). Figure \ref{exp_images/chapter5/edges_experiments} c) illustrates a different idea to connect joints which we refer to as \textit{Range of Motion (ROM) model} since pairs of joints are connected if both lie within the same ROM of a third joint. Further we train an edge configuration as proposed in \cite{arttrack} which we refer to as \textit{Extended model}. 
For completeness, we introduce a \textit{nearly-fully-connected (NFC) model} (Figure \ref{exp_images/chapter5/edges_experiments} e)) which connects most nearby joints. 
We rely on the metric proposed in \cite{deepcut16cvpr} for the estimation of mean average precision (mAP) of all our pose estimation models. Table \ref{table:edge_configs} shows the results, using $\tau_{NSM} = 0.1$.
In all other experiments, we use the Extended model.

	\setlength{\tabcolsep}{4pt}
	\begin{table}
		\centering

		\scalebox{0.75}{\begin{tabular}{lllllllllll}
			\hline\noalign{\smallskip}
			Model & VGG Layers & Trained on  & Head & Shou & Elb  & Wri  & Hip  & Knee & Ankl & Total mAP\\
			\noalign{\smallskip}
			\hline
			\noalign{\smallskip}
	        Standard & 12 & MSCOCO + PoseTrack & 82.9 & 80.3 & 69.9 & \textbf{59.0} & 67.8 & 59.2 & 51.4 & 68.3 \\
            Bypass & 12 & MSCOCO + PoseTrack & \textbf{83.0} & 79.2 & 67.6 & \textbf{59.0} & 66.2 & 61.2 & 53.6 & 68.2 \\
            ROM & 12 & MSCOCO + PoseTrack & 82.0 & 76.2 & 70.3 & 57.9 & 69.3 & 61.7 & 54.1 & 68.3 \\
            \textbf{Extended} & \textbf{12} & \textbf{MSCOCO + PoseTrack} & 80.0 & \textbf{80.8} & \textbf{71.3} & 57.8 & \textbf{72.5} & \textbf{63.3} & \textbf{53.9} & \textbf{69.3} \\
            NFC & 12 & MSCOCO + PoseTrack & 78.3 & 75.8 & 68.3 & 56.9 & 69.2 & 62.1 & 53.5 & 67.1 \\
			\hline
		\end{tabular}}
		\caption{The evaluation of different edge configurations reveals that the {\it Extended} edge configuration performs best compared to the {\it Standard} edge configuration.}
		\label{table:edge_configs}
	\end{table}
	\setlength{\tabcolsep}{1.4pt}

\end{document}